\documentclass{article} 
\usepackage{iclr2022_conference,times}
\usepackage{hyperref}       
\usepackage{url}            
\usepackage{svg}
\usepackage{booktabs}       
\usepackage{bm}
\usepackage{wrapfig} 
\usepackage{amsfonts} 
\usepackage{natbib}
\setcitestyle{square,numbers,sort}
\usepackage{amsthm}
\usepackage{amsmath}
\usepackage{multirow}
\usepackage{microtype}      
\usepackage{subfigure}
\usepackage{longtable}
\usepackage{algorithm}
\usepackage{algorithmic}
\usepackage{svg}
\usepackage{ulem} 
\usepackage{wrapfig}
\usepackage{bbding}
\usepackage{pgfplots}
\usepackage{pgfplotstable}
\pgfplotsset{width=7.5cm,compat=1.13}

\usepackage{amsmath,amsfonts,bm}

\def\eqref#1{equation~\ref{#1}}

\def\1{\bm{1}}

\DeclareMathAlphabet{\mathsfit}{\encodingdefault}{\sfdefault}{m}{sl}
\SetMathAlphabet{\mathsfit}{bold}{\encodingdefault}{\sfdefault}{bx}{n}

\usepackage{hyperref}
\usepackage{url}

\title{M6-10T: A Sharing-Delinking Paradigm for Efficient Multi-Trillion Parameter Pretraining}

\author{%
  Junyang Lin$^{*}$, An Yang$^{*}$, Jinze Bai, Chang Zhou \\
  \textbf{Le Jiang, Xianyan Jia, Ang Wang, Jie Zhang, Yong Li, Wei Lin}\\ 
  \textbf{Jingren Zhou, Hongxia Yang$^{\dagger}$}\\
  Alibaba Group\\
  \texttt{\{junyang.ljy,ya235025,jinze.bjz,ericzhou.zc,jiangle.jl,} \\ 
  \texttt{xianyan.xianyanjia,wangang.wa,wanglin.zj,jiufeng.ly,weilin.lw}\\
  \texttt{jingren.zhou,yang.yhx\}@alibaba-inc.com} \\
}

\iclrfinalcopy 
\begin{document}

\maketitle

\begin{abstract}
Recent expeditious developments in deep learning algorithms, distributed training, and even hardware design for large models have enabled training extreme-scale models, say GPT-3 and Switch Transformer possessing hundreds of billions or even trillions of parameters. 
However, under limited resources, extreme-scale model training that requires enormous amounts of computes and memory footprint suffers from frustratingly low efficiency in model convergence. 
In this paper, we propose a simple training strategy called ``Pseudo-to-Real'' for high-memory-footprint-required large models. 
Pseudo-to-Real is compatible with large models with architecture of sequential layers. 
We demonstrate a practice of pretraining unprecedented $10$-trillion-parameter model, an order of magnitude larger than the state-of-the-art, on solely $512$ GPUs within $10$ days.  
Besides demonstrating the application of Pseudo-to-Real, we also provide a technique, Granular CPU offloading, to manage CPU memory for training large model and maintain high GPU utilities. 
Fast training of extreme-scale models on a decent amount of resources can bring much smaller carbon footprint and contribute to greener AI.


\end{abstract}

\renewcommand{\thefootnote}{\fnsymbol{footnote}}
\footnotetext[1]{Equal contribution.}
\footnotetext[2]{Corresponding author.}
\renewcommand{\thefootnote}{\arabic{footnote}}

\section{Introduction}

The expeditious growing of foundation models on broad data highly contributes to the development of the whole deep learning and artificial intelligence community. 
Foundation models with self-supervised learning on big data have become an emerging paradigm of artificial intelligence systems~\citep{foundation_models}, as they mostly possess high transferability to a wide range of downstream tasks and even multiple modalities. 
The scale of foundation models across domains, including natural language processing, computer vision, and cross-modality representation learning, have been growing tremendously from millions to trillions of parameters~\citep{bert, gpt, gpt2, T5, turing-nlg, megatron, gpt3, dalle, gshard, switch, m6, cpm, cpm-2, m6-t, pangu-alpha, ernie3} thanks to the concurrent advancement in distributed training framework~\citep{megatron_2, zero, zero_offload, zero_infinity, fairscale, whale, deepspeed} and hardware design, and these studies have made a demonstration of the neural scaling law~\citep{scaling_law}. 
However, the training of these transformer-based models incurs high financial costs and even environmental damage due to the massive carbon footprint and thus training extreme-scale models under a decent amount of resources but with high efficiency should be a fundamental goal for both the research and industrial communities to achieve, which promotes the progress of greener AI~\citep{carbon_emission, green_ai}. 

Generally there are two tracks of research in large-scale pretraining, dense models and sparse expert models respectively. 
A typical case of large-scale dense models is GPT-3~\citep{gpt3}. 
It is a 175-billion-parameter transformer model trained with $10,000$ GPUs for months, incurring striking financial and environmental costs. 
Researchers have been looking for methods to training large-scale models with a decent amount of costs. 
Solutions include effective management of memory with gradient and optimizer state partitioning~\citep{zero} or more efficient model parallelism and pipeline parallelism~\citep{megatron, megatron_2, whale}. 
A series of following studies apply those techniques to realize fast training of 10-billion-parameter transformers with hundreds of GPUs in $1-2$ months.~\citep{m6, cpm-2, pangu-alpha, ernie3} 
Sparse expert models with large model capacity are capable of fast training owing to the combination of data parallelism and expert parallelism~\citep{moe, gshard, switch, hash-layers}, and it is even accessible to train a 1-trillion-parameter transformer with no more than $500$ GPUs~\citep{m6-t}. 

Be there as it may, a question emerges in our mind: is it possible to train an extreme-scale model with only a decent amount of resources, e.g., training a 10-trillion-parameter model with $500$ GPUs? 
Such training requires large memory for parameters, including weights, gradients, and even optimizer states. 
Tackling the problem requires the utilization of external memory except for GPU memory, for instance, CPU memory or even NVMe storage~\citep{zero_offload, zero_infinity}. 
These methods resolve the problem of high memory footprint, but instead, their extra cost is low training efficiency caused by the frequent swap in-and-out between memories. 

In this paper, we provide a solution to training large models that require high memory footprint, and we demonstrate a successful practice of pretraining unprecedented extreme-scale models with over $10$ trillion parameters, an order of magnitude larger than the previous state-of-the-arts~\citep{switch, m6-t}. The whole pretraining was conducted on solely $512$ NVIDIA-V100 GPUs and lasts around $10$ days. 
A simple and effective training strategy called ``Pseudo-to-Real'' enables sharing and delinking parameters. 
This training strategy is compatible with architectures built by stacking layers with an identical structure, including dense models like GPT~\citep{gpt, gpt2, gpt3}, BERT~\citep{bert}, or sparse expert models like M6~\citep{m6, m6-t}. 
It is essentially a two-stage training, where in the first stage, we apply cross-layer parameter sharing that requires much less memory footprint for efficient convergence, and in the second, we delink the parameters for better performance. 
It first trains a relatively small model but with a computation graph of a large one with the utilization of cross-layer parameter sharing, and we name it ``Pseudo Giant''. Then it builds a correspondingly large model and delinks the parameters of the shared layer for second-stage model initialization. 
In this way, we achieve fast convergence in the first stage as the training costs much less memory and speeds up with large batches. Parameter sharing that addresses the communication overhead improves training speed as well. 
The second-stage training is responsible for the final convergence for better performance. 

We unlock the secret of pretraining an unprecedented extreme-scale model with over 10 trillion parameters on limited resources of $512$ GPUs. Compared with the previous M6-T on around $500$ GPUs, we do not have a significant increase in computation resources but level up the model scale by an order of magnitude. Besides the application of the ``Pseudo-to-Real'' training strategy, we provide a faster offloading mechanism for both management of CPU memory for parameter storage and utility of GPUs. We successfully train the M6-10T within $10$ days to reach strong performance in log perplexity evaluation and outperform the baseline M6-T. 

Contributions at a glance are below:
\begin{itemize}
    \item We illustrate the training difficulty of extreme-scale models on limited resources and provide a simple but effective solution called ``Pseudo-to-Real''. Upstream and downstream evaluation demonstrates the effectiveness of the strategy. 
    \item We further demonstrate a successful practice of pretraining a 10-trillion-parameter model on $512$ GPUs and reach an outstanding performance within $10$ days. 
\end{itemize}




\section{Related Work}

\paragraph{Large-Scale Pretraining}

In recent years, pretrained language models with growing magnitudes of parameters have been proposed, keeping to raise the validated upper limit of scaling law for model capacity w.r.t the number of parameters~\citep{scaling_law}. Earlier milestones of extreme-large models come from GPT-2~\citep{gpt2} and  Megatron-LM~\citep{megatron}, which demonstrates that scaling the transformer model up to billions of parameters can result in improvement on language modeling benchmarks~\citep{wikitext, lambada}. Turing-NLG \citep{turing-nlg}, as a successor, implements a 17-billion-parameter transformer and achieves further lower perplexity. Similar phenomena are also observed on classification language tasks by T5 model~\citep{T5}. 
The GPT-3~\citep{gpt3} pushes the boundary of model scale to 1,750 billion parameters and demonstrates its striking effectiveness on downstream tasks in even zero-shot settings. 
Furthermore, large-scale pretraining has recently demonstrated success in other fields, including pretraining on other languages or cross-lingual pretraining~\citep{mt5, cpm-2, pangu-alpha, xlm-r}, cross-modal pretraining~\citep{clip, dalle, m6, cogview, ufc-bert} and code generation \citep{codex}. 
Along with the benefit from increasing the model scale, the concern of unaffordable pretraining cost in time, computation resource and energy keeps emerging~\citep{carbon_emission, foundation_models}, resulting in the strong demand for more efficient and greener large-scale pretraining~\citep{green_ai}.


\paragraph{Methods to Train Larger Models Faster}

Our work is about designing algorithm to train extreme-scale models efficiently. 
Researchers have been demonstrating different types of methods to reach this objective. 
To reduce the computational cost during training, some studies introduce  sparsity to the model. 
Mixture-of-Experts (MoE)~\citep{moe} was proposed and was reintroduced in Mesh Tensorflow~\citep{mesh}. 
It shows that MoE with sparsity can significantly improve the model scale efficiently without increasing computation, and the models achieved state-of-the-art performance in language modeling and machine translation. 
GShard~\citep{gshard} extended it to a tremendously large scale of 600B parameters with the sophisticated collaborated design of model architecture and demonstrated its effectiveness across over 100 languages. Similarly, Switch Transformer reached $1.6$ trillion parameters and showed its strong performance on NLU tasks. 
Those models with high sparsity are computationally efficient, and therefore though they possess large model capacity they still can be trained with high efficiency. 
Most other studies still focus on training dense models to validate the scaling law, and thus the emerged problem is the distributed training of large models.  
The most influential distributed framework should be DeepSpeed that proposed ZERO~\citep{zero} that partitions optimizer states and gradients to multiple GPU devices, and ZERO-offload~\citep{zero_offload} as well as ZERO-Infinity~\citep{zero_infinity} can even offload parameters to CPU memory and NVMe storage. 
The memory footprint management makes training extremely large models on limited resources possible. 
However, such offloading mechanisms still have some defects that they may fail to fully utilize the fast hardware. For example, when offloading all parameters to the CPU, the GPU memory can be idle. In this work, we tackle this issue by proposing a granular offloading mechanism that can determine which parameters to be offloaded.

In addition to reducing the amount of computation in a single iteration, another route to speedup training is to reduce the needed iterations for model convergence. \citet{sparse_transformer} and \citet{prenorm_analysis} propose to put forward the layernorm operations in transformer blocks for more stabilized and faster convergence. \citet{lamb} employs a layer-wise adaptive optimizer to enable super-large pretraining batches. \citet{pld} progressively increases the layer dropping rate in a stochastic manner which significantly speedups pretraining.



\section{Approach}
\label{sec:approach}

\begin{figure*}[t]
    \centering
    \includegraphics[scale=0.5]{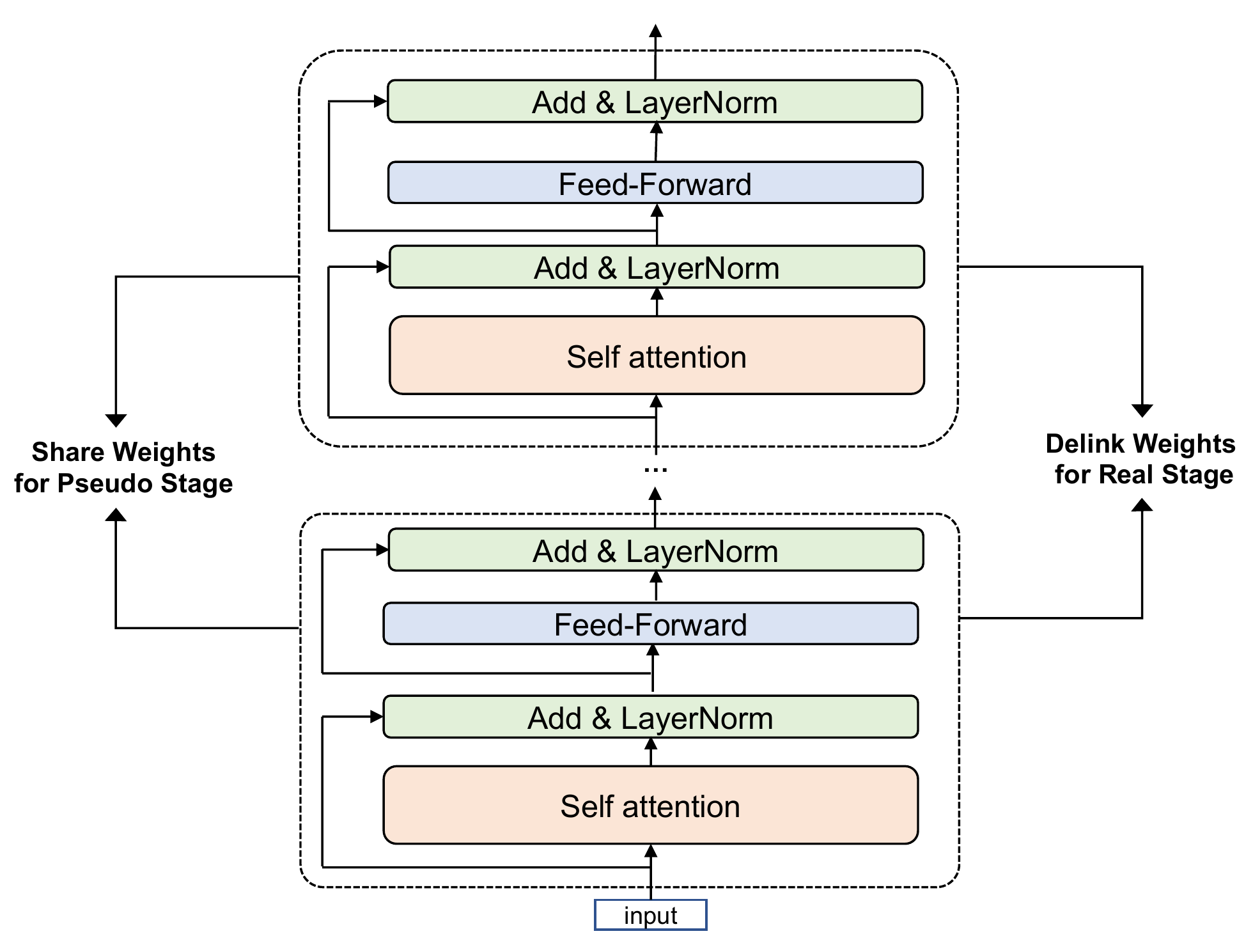}
    \caption{A demonstration of the Pseudo-to-Real training paradigm. It first shares parameters across layers at the Pseudo stage, and then delinks the parameters at the Real stage. }
    \label{fig:pseudo_to_real}
\end{figure*}

This section describes our proposed two-stage training strategy, ``Pseudo-to-Real'', and shows experimental results to validate its effectiveness for training high-memory-footprint-required large models.  

\subsection{Model Architecture}
Choice in model architecture depends on several factors. First, the architecture should contain a sequence of stacking layers, as the sequential structure enables cross-layer parameter sharing. We prefer a simple encoder or decoder architecture, instead of an encoder-decoder framework where there are cross attentions that bring extra parameters and incur difficulties in activation checkpointing. Second, a model of such architecture should be compatible with different types of downstream tasks including understanding and generation, and it is even better that it can be compatible with multiple modalities. Third, as we mention that dense models and sparse expert models are two main tracks of large-scale pretraining, we prefer the model that can flexibly become whether dense or sparse expert models. Therefore, we select M6~\citep{m6} as an option, and we evaluate the effects of our method on M6 of different scales and types. 

M6 is built with stacking transformer layers, which includes self attention and feed-forward neural nets (FFN). For the transformation from dense models to sparse expert models, we should only replace FFN layers with the Mixture-of-Expert (MoE) layers. MoE consists of multiple experts, which are usually FFNs distributed on different devices. A gating network decides the dispatching and combining behaviors of each token and thus tokens can be processed in diverse devices. Such mechanism is a combination of data parallelism and expert parallelism, and thus it is highly efficient though with large model capacity. For the training, to realize the learning of both understanding and generation, the model is trained with text denoising and language modeling on plain text data and with image-based text denoising and image captioning on multimodal data. The model is compatible with different types of downstream tasks and can process information of multiple modalities. 

\subsection{Pseudo-to-Real}

This section demonstrates the details of ``Pseudo-to-Real'' two-stage training strategy that enables fast training of high-memory-footprint-required transformer models. 
The strategy consists of two stages. The first stage trains a model with many fewer parameters but with a large computation graph (``Pseudo Giant''), and the second stage trains a corresponding large model (``Real Giant'') initialized with the delinked weights of the shared layer. Thus we name the strategy ``Pseudo-to-Real'', and the general idea is demonstrated in Figure~\ref{fig:pseudo_to_real}.  

\subsubsection{``Pseudo'' Stage: Layer-wise Parameter Sharing}
The core of ``Pseudo'' stage is to train a Pseudo-Giant that shares parameters across layers. Cross-layer parameter sharing has proved successful in maintaining satisfactory performance while keeping a much smaller amount of parameters. The method was first mentioned in the original Transformer~\citep{transformer}, and \citet{universal_transformer} and \citet{dqe} both illustrated that it is effective for vanilla transformer with encoder-decoder work. \citet{albert} introduced the method to pretraining and proposed a lite BERT with different sharing techniques. Owing to its effectiveness, we introduce it to training an extreme-scale model, and we hypothesize that the first-stage training can gain benefits from cross-parameter sharing as it can address communication overhead and it consumes much less memory footprint. Also, as Pseudo Giant with much fewer parameters is not bounded by memory, it can be trained with large batches for acceleration. 

Suppose we build an M6 model with $L$ layers that share parameters across all layers. The Pseudo Giant though consists of a computation graph of a $L$-layer transformer, its number of weight parameters and optimizer states should be $1/L$ of those of the original one. As to the gradients, we can accumulate the gradients of each layer in the backward computation process, and therefore the amount of gradients becomes $2/L$ of the original one. Such saving in memory enables much faster training with larger batches. Also, it is capable to use fewer resources even due to lower memory consumption. 

This can also be applied to MoE models, as their architecture is stacking transformer layers. However, different from dense models, MoE models partition their weights to all devices and redistribute token representations by sparse activation. This limits the flexible usage of GPU resources. It is available to choose different numbers of GPU devices at different stages in training dense models. In contrast, the restoration of MoE model parameters requires the same number of GPU devices used at the last stage. 
To tackle this problem, we take advantage of the memory efficiency of the Pseudo stage, and distribute more experts to a single GPU and partition experts to more GPUs. 
Suppose we train a model with $512$ experts at each MoE layer. It is possible to train a Pseudo Giant with only $256$ GPUs, where there are $2$ experts on each GPU, and then train a Real Giant on $512$ GPUs where there is only $1$ expert on each GPU.

\subsubsection{``Real'' Stage: Delinking the Shared Parameters}
We name a large model without cross-layer parameter sharing ``Real Giant'', in comparison with Pseudo Giant. Both Pseudo Giant and Real Giant share a computation graph, but they have different numbers of parameters. In this work, we discover how to build a connection between the two models. 
Given a Pseudo Giant fully trained until convergence, we apply the delinking of cross-layer shared parameters to accelerate Real Giant training. 
There is no need to train a large model from scratch. The model can start its convergence from low perplexity.

Embedding initialization can be directly restored, but the layer weights should be treated specially. In practice, there is only one layer of weights $\theta_{shared}$ in Pseudo Giant, and there are $L$ layers of weights $\{\theta_{1}, \theta_{2}, \cdots, \theta_{L}\}$ in Real Giant. Thanks to their identical structure, each layer of Real Giant can be initialized with $\theta_{shared}$. Without further training, this model is equivalent to a fully-trained Pseudo Giant. 

This extremely simple training strategy is highly beneficial for the high-memory-footprint-required large models, especially extreme-scale models like the 10-trillion-parameter M6. While the first stage of training saves much time for faster convergence, we can use a decent amount of computational resources in this stage as lower efficiency in this stage becomes acceptable. Therefore, in the practice of training an extreme-scale M6, we apply CPU offloading to utilize CPU memory. Therefore, we can use a limited amount of resources, e.g., $512$ GPUs, to train an unprecedented 10-trillion-parameter model efficiently, which is an order of magnitude larger than the state-of-the-arts. 


\subsubsection{Timing for Switching}
A question naturally emerges: when should we switch from the Pseudo stage to the Real stage?  As mentioned above, the greatest advantage of Pseudo stage for training is the significantly faster convergence speed. Yet the performance of Pseudo Giant is bounded by its limitation in the number of parameters. Training Pseudo Giant until its final convergence apparently incurs much waste of time. 

In practice, we present a simple strategy to determine the training step to switch from Pseudo to Real based on the convergence speed. 
During the training of the Pseudo stage, we evaluate a training step in a fixed interval by attempting to transfer it into the Real stage and training for a small while (e.g., $30$ minutes).
After that, we will revert the model parameters to the evaluated step and continue the training of the Pseudo stage for the same training time as the Real stage.
If the decreasing speed of loss in the Real stage surpasses that of the Pseudo stage, we determine the evaluated training step as the best switching point for the next-stage training.


\subsection{Experiments}
In this section, we provide a series of experiments to evaluate the effectiveness of the training strategy by observing the models' upstream and downstream performance. 

\subsubsection{Experimental Setup}
We conduct experiments for pretraining and finetuning to analyze model competence in upstream and downstream tasks. 
Following the classical data setup for pretraining and finetuning, we pretrain the model on BookCorpus~\citep{bookcorpus} and English Wikipedia~\citep{bert}, which are corpora with around 16GB of plain texts.

We validate the effectiveness of ``Pseudo-to-Real'' training strategy by implementing a medium-size M6 model to conduct extensive analyses on both upstream and downstream performance. 
To satisfy the requirements of high memory footprint where the two-staged training can make difference in training efficiency, we conduct experiments on a 1.4B-parameter model, the largest model trained on a single NVIDIA V100-32GB~\citep{zero_offload}. The corresponding Pseudo Giant is a model with the same computation graph but sharing parameters across layers. 

Following \citet{gpt} and \citet{bart}, we use a vocabulary of around $50,000$ subwords. 
Each sample consists of sentences from an identical passage, and we use a sequence length of $512$ and correspondingly truncate or pad the sequence. 
We build an M6 model with $36$ layers of transformer, whose hidden size is $1024$ and intermediate size is $16384$. 
As our evaluation is conducted on plain text data, we adopt the two tasks, text denoising and language modeling, for pretraining.
We name the one with cross-layer parameter sharing ``Pseudo'' and the one without it ``Real'' in Table~\ref{tab:model_config}. The total number of parameters of Real Giant is around $1.4$ billion and that of its corresponding Pseudo Giant is around $90$ million. 
We use ``P2R'' referring to ``Pseudo-to-Real'' to represent the model trained with both ``Pseudo'' and ``Real'' stages. 
``Pseudo'' and ``Real'' refer to the models that are pretrained from scratch, in contrast with ``P2R''. 
Furthermore, we build an M6 model of $358$M parameters as a baseline. It has a smaller intermediate size of $4096$ and consists of $24$ layers. We name it ``Base'' in Table~\ref{tab:model_config}.

Following the common practice in pretraining~\citep{bert, roberta}, we apply AdamW optimizer~\citep{adamw} for optimization.
To determine the most suitable learning rate of the two stages for fast convergence, we have made some preliminary tests and finally used the peak learning rate of $2e-4$ for ``Pseudo'' stage and $8e-5$ for ``Real'' stage, respectively.
We use a cosine decaying mechanism for learning rate scheduling and a warm-up ratio of $0.001$.
For both Pseudo and Real Giant, we train them with a total batch size of $6,144$. In practice, we use a micro-batch size of $32$ and a gradient accumulation step of $4$, and we train our models on $48$ NVIDIA-V100 GPU devices. 
We create two setups for better comparison. The one is ``limited budget'', where models have been trained for the same duration of time. Correspondingly, ``Pseudo'' has been trained for around $25,000$ steps, Real has been trained for around $4,700$ steps, and ``P2R'' has been trained for $12,000$ steps. 
The other is ``training for longer'', which demonstrates a wall-time performance of M6-1B for better comparison. 

\subsubsection{Downstream Evaluation}
We compare the model performance by the downstream evaluation on language modeling and text summarization tasks, which cover capabilities including language modeling and generation. To be more specific, the downstream datasets include:
\begin{itemize}
    \item \textbf{WikiText103}: a classical language modeling evaluation benchmark dataset that consists of Wikipedia articles. We evaluate the Perplexity (PPL) of the pretrained model, which is a per-token exponential cross-entropy loss that reflects probability distribution over texts. 
    \item \textbf{Gigaword}: a dataset for summarization which consists of around 3.8M articles and summaries, and an effective benchmark to evaluate the model's capability in text generation for abstractive summarization. 
\end{itemize}

\begin{table}
\centering
  \caption{Model refers to the types of model. $d_{model}$ and $d_{ff}$ refer to the hidden size and intermediate size. $L$ refers to the number of layers in the computation graph, and $l$ refers to the number of transformer layers with parameters. \#Heads refers to the number of heads in self attention. \#Params refers to the total number of model parameters. We also report their training speed on $48$ GPU devices by the number of consumed samples per second. } 
  \label{tab:model_config}
  \begin{tabular}{ccccccc}
    \toprule
    Model & $d_{model}$ & $d_{ff}$ & l/L & \#Heads &  \#Params  & Speed \\
    \midrule
    Base & 1024  & 4096  & 24/24 & 16  & 350M     &  650\\
    Pseudo & 1024  & 16384  & 1/36  & 16  & 1.4B & 248  \\
    Real & 1024  & 16384  & 36/36  & 16  & 1.4B & 48 \\
    \bottomrule
  \end{tabular}
\end{table}

For better comparison, we also present the other pretrained models as baselines to demonstrate that the models can achieve strong performance in downstream tasks. Furthermore, we also compare the model quality on the time basis to reflect the advantage of Pseudo-to-Real training strategy. 

To view their training efficiency, we focus on their training speed on the condition that we attempt to exhaust the GPU utility. 
Table~\ref{tab:model_config} demonstrates the training speed of the models. We report their training speed on $48$ NVIDIA-V100 GPU devices with their consumed samples per second.  
Obviously training Real Giant model from scratch is highly time-consuming, and Pseudo Giant training has an advantage of around $5$ times of convergence speed over Real Giant training. 

\begin{wrapfigure}{r}{7.0cm}
  \vspace{-15pt}    
  \includegraphics[width=7.0cm]{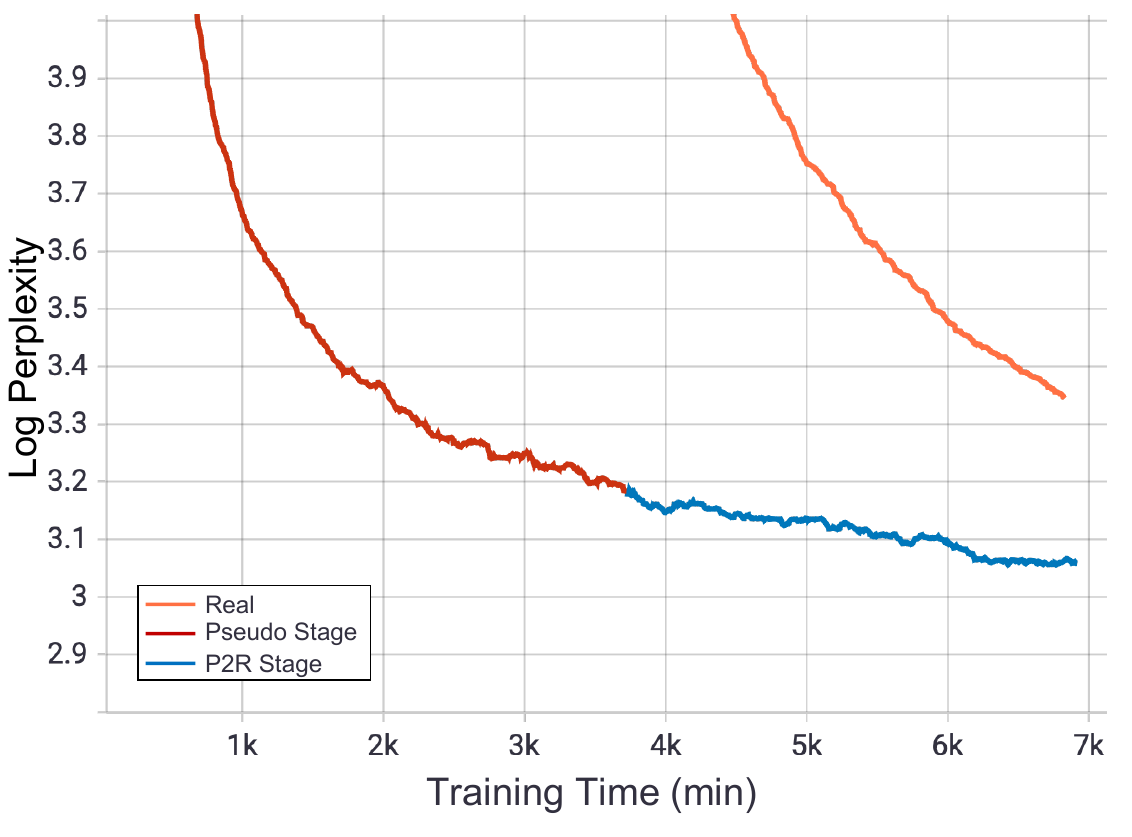}\\
  \vspace{-15pt}    
  \caption{Comparison of pretraining language modeling loss of M6-1B P2R and Real on time-basis.}
  \vspace{-5pt}    
  \label{fig:loss_convergence}
\end{wrapfigure}

We also observe the loss convergence of both P2R and Real Giant trained from scratch. 
In Figure~\ref{fig:loss_convergence} we present the development of pretraining language modeling loss, which is the log perplexity, on the time basis. 
The log perplexity of P2R decreases much faster than that of the Real Giant, with an advantage larger than $0.3$.

\begin{table}
\centering
  \caption{Experimental results on downstream task evaluation. ``\#Params'' refers to the number of parameters. We report the PPL evaluation on WikiText-103 and the ROUGE1, ROUGE-2, and ROUGE-L evaluation on Gigaword. } 
  \label{tab:results}
  \begin{tabular}{cccc}
    \toprule
    Model       & \#Params       & WikiText-103  & Gigaword        \\
    \midrule
    Megatron-LM & 350M          & 16.69          & -               \\
    UniLM       & 340M             & -          & 38.5/19.5/35.4    \\
    M6          & 350M          &   16.59          & 38.8/20.1/36.0      \\
    \midrule
    \multicolumn{4}{c}{\small{\textbf{limited budget}}} \\
    M6-1B (Pseudo)      & 90M          & 56.79          & 36.8/17.9/34.1            \\
    M6-1B (Real)      & 1.4B          & 26.80          & 36.9/18.2/34.2          \\
    M6-1B (P2R)        & 90M/1.4B          & 23.60     & 37.3/18.3/34.5         \\
    \midrule
    \multicolumn{4}{c}{\small{\textbf{training for longer}}} \\
    M6-1B        & 90M/1.4B          & 16.52     & 38.0/19.2/35.4          \\
    \bottomrule
  \end{tabular}
\end{table}


We add other strong baseline pretrained models for the two tasks for comparison. 
M6-base model matches the performance of pretrianed models, including UniLM, Megatron-LM GPT, etc.~\citep{unilm, megatron}. 
Experimental results are consistent with our hypothesis that Pseudo-to-Real training can speed up training effectively. 
In the setup of limited budget, the M6 model trained with Pseudo-to-Real can outperform the Pseudo Giant and Real Giant in both language modeling and text generation. 
We also add an M6-1B model trained for a longer time to show its wall-time performance on downstream tasks. 
However, we did not reach the best performance in Gigaword. 
As it is not related to the training strategy, we leave this issue to future work for further discussion about how to finetune large models on downstream tasks. 

\section{Towards a 10-Trillion-Parameter Model}

Previously, training large-scale models brings tons of challenges to the collaboration algorithm design, distributed training, as well as hardware design, etc. Training a GPT-3 of $175$B parameters with a combination of data parallelism on over $500$GB of data should cost around $300$ GPU-years. Later with the emergence of partitioning on optimizer states, gradients, and even weights, GPU memory can be fully utilized without performance degradation. Now we can even use the CPU memory or even NVMe storage to store the parameters, but we have to bear the costs of efficiency. Therefore, we attempt to tackle the difficulty of extreme-scale model training from the perspective of algorithm design and thus we apply the aforementioned Pseudo-to-Real training strategy to train an extreme-scale model.

\subsection{Model Setup}
We design a 10-trillion-parameter M6 model with the combination of existing methods and proposed strategies to demonstrate a case of how to train an extreme-scale model efficiently. In comparison with the previous studies of trillion-parameter models~\citep{switch, m6-t}, this one is almost 10 times larger. To efficiently utilize the memory, we adopt Mixture-of-Experts and we replace every FFN layer with the memory-efficient MoE layers. Notably, we remove the auxiliary loss that consumes memory and demonstrates little effects on model quality, and we follow \citet{m6-t} to apply expert prototyping for improved model quality and training stability. 
To be more specific, the hidden size $d_{model}$ is $1024$ and the intermediate size $d_{ff}$ is $9984$. The number of model layers is $48$. For each MoE layer, there are $10240$ experts distributed on multiple devices. We use $80$ prototypes of experts based on our experience in preliminary experiments. The training batch size per GPU device is $8$. 
We implement models on EFLOPs, a distributed training platform with an advanced server architecture and a new network topology~\citep{eflops}.
Specifically our models are trained on a cluster of 8-GPU workers connected by RDMA networks with a bandwidth of 100Gb. The CPU memory of each worker is around 750GB. 
The expert distribution and Granular Offloading strategies are supported by Whale framework\cite{whale}. 

\subsection{Granular CPU offloading}
\begin{figure*}[t]
    \centering
    \includegraphics[scale=0.6]{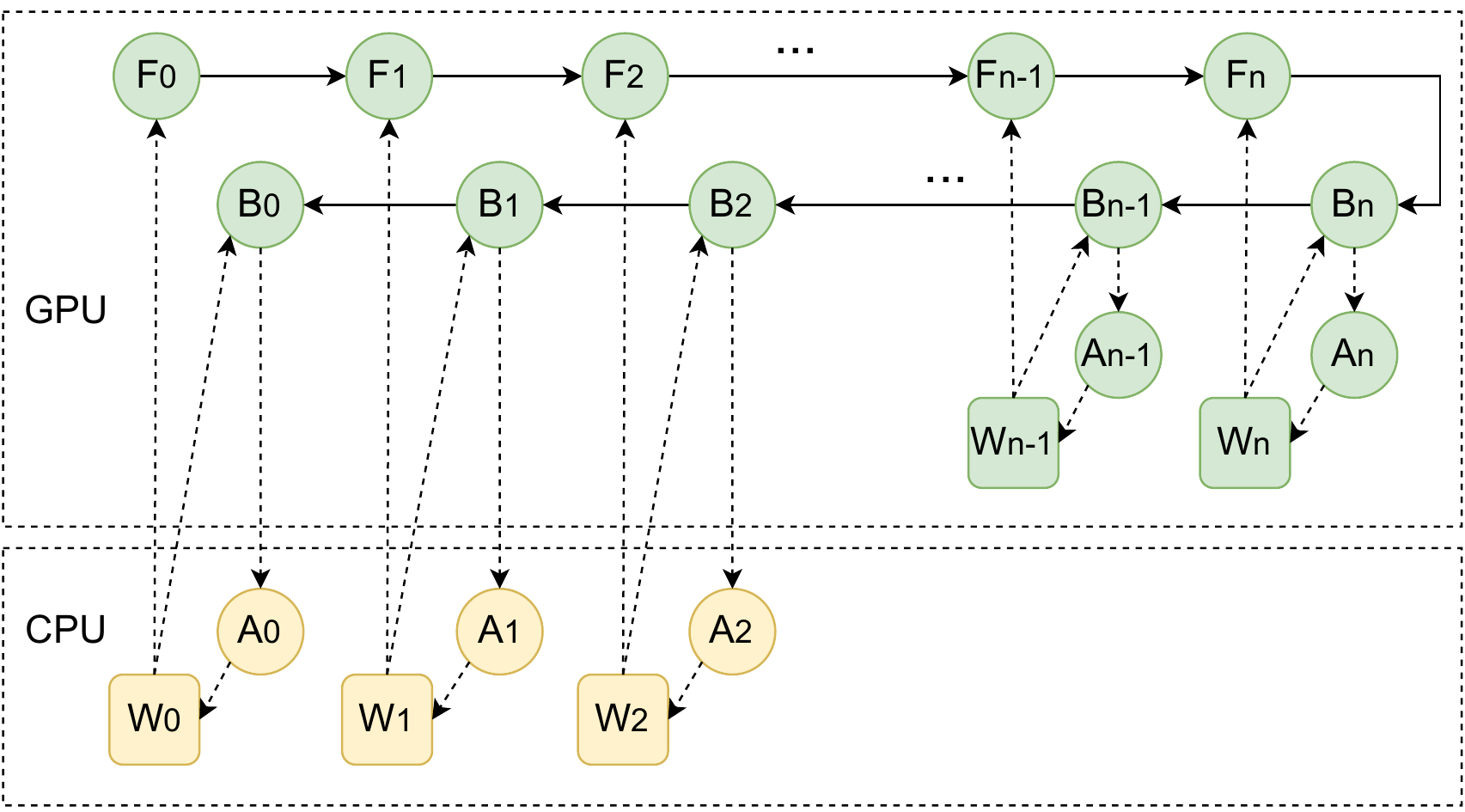}
    \caption{A demonstration of Granular CPU Offloading mechanism. }
    \label{fig:cpu_offload}
\end{figure*}
To utilize CPU memory for larger models with fewer resources, we apply the Granular CPU offloading which has a higher efficiency compared with the conventional CPU offloading. 
Previously we note that conventional offloading mechanisms offload all parameters, which may fail to effectively utilize GPU. 
To improve the efficiency of CPU offloading, we propose a new CPU offloading mechanism called Granular CPU offloading. 
The training process is composed of phases including ``Forward (Fn)'', ``Backward(Bn)'' and ``Apply (An)''. Offloading all model parameters to CPU in Fn and Bn requires loading parameters from CPU to GPU memory twice. 
Activation checkpointing that brings recomputation needs the parameters loaded in Bn. 
An requires the gradients offloaded from GPU memory to CPU memory. Assume the model parameter size is $\mathbf{W}$, the above processes bring parameter movement of size $\mathbf{4*W}$.

In offloading, PCIE is the bottleneck of the whole training process.
We observe that when offloading all parameters with recomputation, the GPU memory is idle. We can fill up the GPU memory by selective offloading, leaving part of the model in GPU memory. In this way, the model can be accelerated by reducing across-device memory copy. 
In our preliminary experiment, with the setting of training a 48-layer 78B-parameter M6 model on 8 NVIDIA-V100 GPU devices, the step-time costs 89 seconds when fully offloading the parameters into CPU memory.
In comparison, granularly offloading the first 24 layers into CPU memory and leaving the remaining 24 layers on GPU reduces the step-time to only 45 seconds.
The significant difference in training step-time indicates that the time-cost of parameter movement between CPU and GPU will dominate the training step-time when offloading is employed, thus the granularity of offloading and the utilization of GPU should be seriously considered in extreme-scale pretraining.
In addition, offloading the whole model can result in OOM error in the CPU when the model is extremely large.

With Granular CPU offloading, we successfully implement a 10-trillion-parameter M6 model on solely 512 NVIDIA-V100 GPUs. 
Furthermore, at the Pseudo stage, we can train a Pseudo Giant with a computation graph of 10 trillion parameters only with 256 GPU devices without the utilization of CPU memory for offloading. 
Thus in our practice, we train a Pseudo Giant with only 256 GPUs, and then partition the experts and redistribute them to 512 GPUs. 
This saves the usage of GPU resources, which is more resource-efficient and also more environmentally friendly.

\subsection{Analysis}

\begin{figure*}[t] \centering
	\subfigure[M6-10T P2R vs. Real on time-basis] {\label{fig:real_vs_p2r_bytime}
		\includegraphics[scale=0.52]{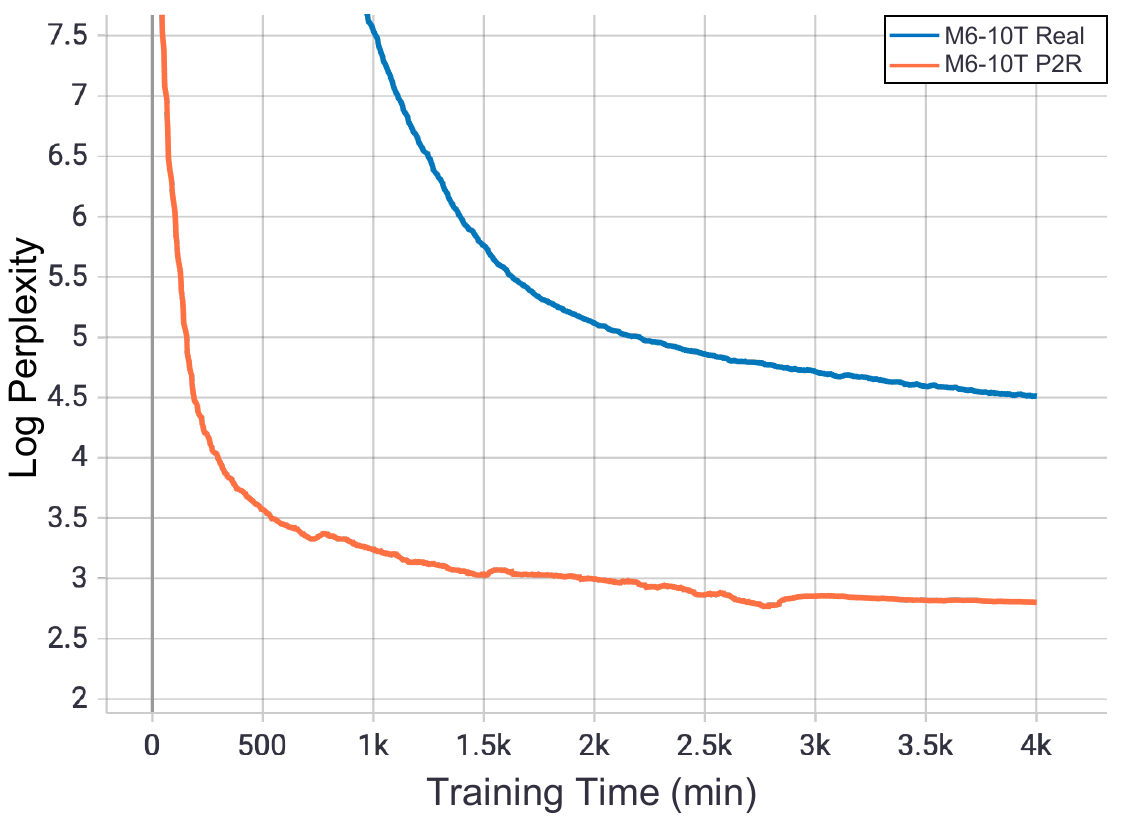}  
	}
	\subfigure[M6-10T P2R vs. M6-T on sample-basis] {\label{fig:1T_vs_10T_bysample}
		\includegraphics[scale=0.52]{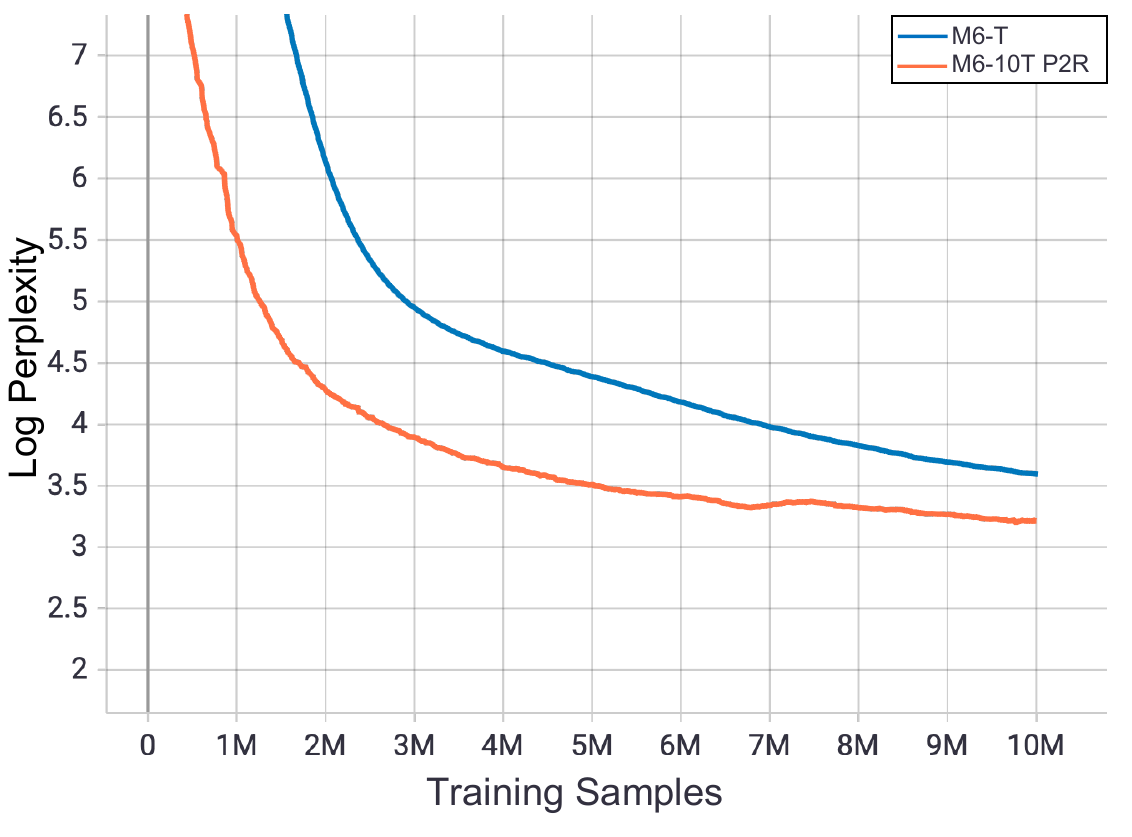}  
	}
	\caption{\label{fig:10T_exp}The log perplexity of M6-10T Pesudo-to-Real (P2R) compared with baselines on time-basis and sample-basis, respectively. (a) Compared with M6-10T Real on time-basis, M6-10T P2R converges much faster due to the significant reduction of time-cost on offloading in the Pseudo stage. (b) Compared with 1T-parameter M6-T model on sample-basis, though undergoing the Pseudo stage which limits the model capacity, M6-10T still has an advantage in training sample efficiency.}
\end{figure*}

\paragraph{Training Efficiency}

We pretrain the M6-10T with Pseudo-to-Real training strategy for around $10$ days, and we additionally train an M6-10T from scratch without the strategy for around $3$ days for comparison. 
The Real stage training can be facilitated without Out-of-Memory errors with the help of Granular CPU offloading, but its step-time is only around $180$s. 
In contrast, the step-time of Pseudo stage is only $14$s without the cost of offloading, which greatly boosts the training efficiency of M6-10T P2R.
The M6-10T with P2R has been trained for $15k$ steps, but the one from scratch has been trained for solely $1.3k$ steps. 
We record the log perplexity of both models trained on the M6-Corpus on the time basis in Figure~\ref{fig:real_vs_p2r_bytime}. 
Results show that within the same length of time M6-10T with P2R can outperform the one trained from scratch by a large margin. 

\paragraph{Convergence Analysis}
We have trained M6-10T with Pseudo-to-Real strategy for around $10$ days, and the model converges to a low level of log perplexity based on the upstream evaluation on the M6-Corpus. 
For better comparison, we also show the convergence performance of the 1T-parameter M6-T model proposed in the previous work. 
As shown in Figure~\ref{fig:1T_vs_10T_bysample}, the observation is consistent with our intuition that the model with a larger capacity can converge faster on the sample basis, and it should achieve the best performance on language modeling. 
What leaves open is whether it can positively lead to better downstream performance concerning different types of downstream tasks. 
Finetuning extremely large models should be difficult and there is still much room for us to discover the potential of extreme-scale models. 
However, the contribution of this work is leveling up their training efficiency, which can be regarded as an initial step to investigate the secrets of super large models.

\section{Conclusion and Future Work}
Pseudo-to-Real training strategy is a simple and effective way to train large-scale models that are highly memory consuming, and it is also essential to training extremely large models with limited resources with significantly higher training efficiency. 
We unlock pretraining unprecedented extreme-scale models with 10 trillion parameters with limited resources of 512 GPUs in $10$ days. Besides the application of Pseudo-to-Real training strategy, we further provide Granular CPU offloading to enhance GPU utility while breaking the GPU memory wall with a cost in efficiency. 
The advances take a leap towards extreme-scale model training beyond implementation on limited resources. 
With only a few GPU cards, training large models with tens or hundreds of parameters has become accessible to many researchers. 
We believe that this can motivate low carbon dioxide production and encourages the progress of green AI.

\section*{Ethics Statement}
This work is highly concerned with large-scale language models and multimodal pretrained models. These models have been pretrained on broad data of plain texts and image-text pairs, which might contain harmful information, such as hate speech, terrorism, pornography, etc. 
We have put much efforts to remove these kinds of data in our datasets by quality evaluation on texts and images. 
However, this problem cannot be eliminated and ignored, and it is common in the pretraining community. 
For those models that are not trained on commonly-used public datasets, we will carefully release the model checkpoints before careful evaluation, and also limit the access to avoid misconduct.

\section*{Reproducibility Statement}
This work is generally reproducible. Following the description in Section~\ref{sec:approach}, researchers can easily implement the training strategy on the codebases for pretraining, including Huggingface Transformer~\footnote{\url{https://github.com/huggingface/transformers}}, Fairseq~\footnote{\url{https://github.com/pytorch/fairseq}}.

\bibliography{iclr2022_conference}

\begin{thebibliography}{52}
\providecommand{\natexlab}[1]{#1}
\providecommand{\url}[1]{\texttt{#1}}
\expandafter\ifx\csname urlstyle\endcsname\relax
  \providecommand{\doi}[1]{doi: #1}\else
  \providecommand{\doi}{doi: \begingroup \urlstyle{rm}\Url}\fi

\bibitem[Bai et~al.(2019)Bai, Kolter, and Koltun]{dqe}
Shaojie Bai, J.~Zico Kolter, and Vladlen Koltun.
\newblock Deep equilibrium models.
\newblock In \emph{Advances in Neural Information Processing Systems 32: Annual
  Conference on Neural Information Processing Systems 2019, NeurIPS 2019}, pp.\
   688--699, 2019.

\bibitem[Baines et~al.(2021)Baines, Bhosale, Caggiano, Goyal, Goyal, Ott,
  Lefaudeux, Liptchinsky, Rabbat, Sheiffer, Sridhar, and Xu]{fairscale}
Mandeep Baines, Shruti Bhosale, Vittorio Caggiano, Naman Goyal, Siddharth
  Goyal, Myle Ott, Benjamin Lefaudeux, Vitaliy Liptchinsky, Mike Rabbat, Sam
  Sheiffer, Anjali Sridhar, and Min Xu.
\newblock Fairscale: A general purpose modular pytorch library for high
  performance and large scale training.
\newblock \url{https://github.com/facebookresearch/fairscale}, 2021.

\bibitem[Bommasani et~al.(2021)Bommasani, Hudson, Adeli, Altman, Arora, von
  Arx, Bernstein, Bohg, Bosselut, Brunskill, et~al.]{foundation_models}
Rishi Bommasani, Drew~A Hudson, Ehsan Adeli, Russ Altman, Simran Arora, Sydney
  von Arx, Michael~S Bernstein, Jeannette Bohg, Antoine Bosselut, Emma
  Brunskill, et~al.
\newblock On the opportunities and risks of foundation models.
\newblock \emph{arXiv preprint arXiv:2108.07258}, 2021.

\bibitem[Brown et~al.(2020)Brown, Mann, Ryder, Subbiah, Kaplan, Dhariwal,
  Neelakantan, Shyam, Sastry, Askell, Agarwal, Herbert{-}Voss, Krueger,
  Henighan, Child, Ramesh, Ziegler, Wu, Winter, Hesse, Chen, Sigler, Litwin,
  Gray, Chess, Clark, Berner, McCandlish, Radford, Sutskever, and Amodei]{gpt3}
Tom~B. Brown, Benjamin Mann, Nick Ryder, Melanie Subbiah, Jared Kaplan,
  Prafulla Dhariwal, Arvind Neelakantan, Pranav Shyam, Girish Sastry, Amanda
  Askell, Sandhini Agarwal, Ariel Herbert{-}Voss, Gretchen Krueger, Tom
  Henighan, Rewon Child, Aditya Ramesh, Daniel~M. Ziegler, Jeffrey Wu, Clemens
  Winter, Christopher Hesse, Mark Chen, Eric Sigler, Mateusz Litwin, Scott
  Gray, Benjamin Chess, Jack Clark, Christopher Berner, Sam McCandlish, Alec
  Radford, Ilya Sutskever, and Dario Amodei.
\newblock Language models are few-shot learners.
\newblock In Hugo Larochelle, Marc'Aurelio Ranzato, Raia Hadsell,
  Maria{-}Florina Balcan, and Hsuan{-}Tien Lin (eds.), \emph{Advances in Neural
  Information Processing Systems 33: Annual Conference on Neural Information
  Processing Systems 2020, NeurIPS 2020, December 6-12, 2020, virtual}, 2020.

\bibitem[Chen et~al.(2021)Chen, Tworek, Jun, Yuan, Ponde, Kaplan, Edwards,
  Burda, Joseph, Brockman, et~al.]{codex}
Mark Chen, Jerry Tworek, Heewoo Jun, Qiming Yuan, Henrique Ponde, Jared Kaplan,
  Harri Edwards, Yura Burda, Nicholas Joseph, Greg Brockman, et~al.
\newblock Evaluating large language models trained on code.
\newblock \emph{arXiv preprint arXiv:2107.03374}, 2021.

\bibitem[Child et~al.(2019)Child, Gray, Radford, and
  Sutskever]{sparse_transformer}
Rewon Child, Scott Gray, Alec Radford, and Ilya Sutskever.
\newblock Generating long sequences with sparse transformers.
\newblock \emph{arXiv preprint arXiv:1904.10509}, 2019.

\bibitem[Conneau et~al.(2020)Conneau, Khandelwal, Goyal, Chaudhary, Wenzek,
  Guzm{\'{a}}n, Grave, Ott, Zettlemoyer, and Stoyanov]{xlm-r}
Alexis Conneau, Kartikay Khandelwal, Naman Goyal, Vishrav Chaudhary, Guillaume
  Wenzek, Francisco Guzm{\'{a}}n, Edouard Grave, Myle Ott, Luke Zettlemoyer,
  and Veselin Stoyanov.
\newblock Unsupervised cross-lingual representation learning at scale.
\newblock In \emph{Proceedings of the 58th Annual Meeting of the Association
  for Computational Linguistics, {ACL} 2020}, pp.\  8440--8451, 2020.

\bibitem[Dehghani et~al.(2019)Dehghani, Gouws, Vinyals, Uszkoreit, and
  Kaiser]{universal_transformer}
Mostafa Dehghani, Stephan Gouws, Oriol Vinyals, Jakob Uszkoreit, and Lukasz
  Kaiser.
\newblock Universal transformers.
\newblock In \emph{7th International Conference on Learning Representations,
  {ICLR} 2019, New Orleans, LA, USA, May 6-9, 2019}. OpenReview.net, 2019.

\bibitem[Devlin et~al.(2019)Devlin, Chang, Lee, and Toutanova]{bert}
Jacob Devlin, Ming-Wei Chang, Kenton Lee, and Kristina Toutanova.
\newblock {BERT}: Pre-training of deep bidirectional transformers for language
  understanding.
\newblock In \emph{Proceedings of the 2019 Conference of the North {A}merican
  Chapter of the Association for Computational Linguistics: Human Language
  Technologies, Volume 1 (Long and Short Papers)}, pp.\  4171--4186,
  Minneapolis, Minnesota, 2019. Association for Computational Linguistics.

\bibitem[Ding et~al.(2021)Ding, Yang, Hong, Zheng, Zhou, Yin, Lin, Zou, Shao,
  Yang, and Tang]{cogview}
Ming Ding, Zhuoyi Yang, Wenyi Hong, Wendi Zheng, Chang Zhou, Da~Yin, Junyang
  Lin, Xu~Zou, Zhou Shao, Hongxia Yang, and Jie Tang.
\newblock Cogview: Mastering text-to-image generation via transformers.
\newblock \emph{CoRR}, abs/2105.13290, 2021.

\bibitem[Dong et~al.(2020)Dong, Cao, Zhang, Ye, Wang, Feng, Zhao, Liu, Song,
  Peng, Guo, Jiang, Tang, Du, Zhang, Pan, and Xie]{eflops}
Jianbo Dong, Zheng Cao, Tao Zhang, Jianxi Ye, Shaochuang Wang, Fei Feng,
  Li~Zhao, Xiaoyong Liu, Liuyihan Song, Liwei Peng, Yiqun Guo, Xiaowei Jiang,
  Lingbo Tang, Yin Du, Yingya Zhang, Pan Pan, and Yuan Xie.
\newblock {EFLOPS:} algorithm and system co-design for a high performance
  distributed training platform.
\newblock In \emph{{IEEE} International Symposium on High Performance Computer
  Architecture, {HPCA} 2020}, pp.\  610--622, 2020.

\bibitem[Dong et~al.(2019)Dong, Yang, Wang, Wei, Liu, Wang, Gao, Zhou, and
  Hon]{unilm}
Li~Dong, Nan Yang, Wenhui Wang, Furu Wei, Xiaodong Liu, Yu~Wang, Jianfeng Gao,
  Ming Zhou, and Hsiao{-}Wuen Hon.
\newblock Unified language model pre-training for natural language
  understanding and generation.
\newblock In Hanna~M. Wallach, Hugo Larochelle, Alina Beygelzimer, Florence
  d'Alch{\'{e}}{-}Buc, Emily~B. Fox, and Roman Garnett (eds.), \emph{Advances
  in Neural Information Processing Systems 32: Annual Conference on Neural
  Information Processing Systems 2019, NeurIPS 2019, December 8-14, 2019,
  Vancouver, BC, Canada}, pp.\  13042--13054, 2019.

\bibitem[Fedus et~al.(2021)Fedus, Zoph, and Shazeer]{switch}
William Fedus, Barret Zoph, and Noam Shazeer.
\newblock Switch transformers: Scaling to trillion parameter models with simple
  and efficient sparsity.
\newblock \emph{arXiv preprint arXiv:2101.03961}, 2021.

\bibitem[Jia et~al.(2020)Jia, Jiang, Wang, Zhang, Li, Xiao, Li, Zheng, Liu, and
  Lin]{whale}
Xianyan Jia, Le~Jiang, Ang Wang, Jie Zhang, Xinyuan Li, Wencong Xiao, Yong Li,
  Zhen Zheng, Xiaoyong Liu, and Wei Lin.
\newblock Whale: Scaling deep learning model training to the trillions.
\newblock \emph{arXiv preprint arXiv:2011.09208}, 2020.

\bibitem[Kaplan et~al.(2020)Kaplan, McCandlish, Henighan, Brown, Chess, Child,
  Gray, Radford, Wu, and Amodei]{scaling_law}
Jared Kaplan, Sam McCandlish, Tom Henighan, Tom~B Brown, Benjamin Chess, Rewon
  Child, Scott Gray, Alec Radford, Jeffrey Wu, and Dario Amodei.
\newblock Scaling laws for neural language models.
\newblock \emph{arXiv preprint arXiv:2001.08361}, 2020.

\bibitem[Lan et~al.(2020)Lan, Chen, Goodman, Gimpel, Sharma, and
  Soricut]{albert}
Zhenzhong Lan, Mingda Chen, Sebastian Goodman, Kevin Gimpel, Piyush Sharma, and
  Radu Soricut.
\newblock {ALBERT:} {A} lite {BERT} for self-supervised learning of language
  representations.
\newblock In \emph{8th International Conference on Learning Representations,
  {ICLR} 2020, Addis Ababa, Ethiopia, April 26-30, 2020}. OpenReview.net, 2020.

\bibitem[Lepikhin et~al.(2020)Lepikhin, Lee, Xu, Chen, Firat, Huang, Krikun,
  Shazeer, and Chen]{gshard}
Dmitry Lepikhin, HyoukJoong Lee, Yuanzhong Xu, Dehao Chen, Orhan Firat, Yanping
  Huang, Maxim Krikun, Noam Shazeer, and Zhifeng Chen.
\newblock Gshard: Scaling giant models with conditional computation and
  automatic sharding.
\newblock \emph{arXiv preprint arXiv:2006.16668}, 2020.

\bibitem[Lewis et~al.(2020)Lewis, Liu, Goyal, Ghazvininejad, Mohamed, Levy,
  Stoyanov, and Zettlemoyer]{bart}
Mike Lewis, Yinhan Liu, Naman Goyal, Marjan Ghazvininejad, Abdelrahman Mohamed,
  Omer Levy, Veselin Stoyanov, and Luke Zettlemoyer.
\newblock {BART}: Denoising sequence-to-sequence pre-training for natural
  language generation, translation, and comprehension.
\newblock In \emph{Proceedings of the 58th Annual Meeting of the Association
  for Computational Linguistics}, pp.\  7871--7880, Online, 2020. Association
  for Computational Linguistics.

\bibitem[Lin et~al.(2021)Lin, Men, Yang, Zhou, Ding, Zhang, Wang, Wang, Jiang,
  Jia, et~al.]{m6}
Junyang Lin, Rui Men, An~Yang, Chang Zhou, Ming Ding, Yichang Zhang, Peng Wang,
  Ang Wang, Le~Jiang, Xianyan Jia, et~al.
\newblock M6: A chinese multimodal pretrainer.
\newblock \emph{arXiv preprint arXiv:2103.00823}, 2021.

\bibitem[Liu et~al.(2019)Liu, Ott, Goyal, Du, Joshi, Chen, Levy, Lewis,
  Zettlemoyer, and Stoyanov]{roberta}
Yinhan Liu, Myle Ott, Naman Goyal, Jingfei Du, Mandar Joshi, Danqi Chen, Omer
  Levy, Mike Lewis, Luke Zettlemoyer, and Veselin Stoyanov.
\newblock Roberta: {A} robustly optimized {BERT} pretraining approach.
\newblock \emph{arXiv preprint arXiv:1907.11692}, 2019.

\bibitem[Loshchilov \& Hutter(2019)Loshchilov and Hutter]{adamw}
Ilya Loshchilov and Frank Hutter.
\newblock Decoupled weight decay regularization.
\newblock In \emph{7th International Conference on Learning Representations,
  {ICLR} 2019, New Orleans, LA, USA, May 6-9, 2019}. OpenReview.net, 2019.

\bibitem[Merity et~al.(2017)Merity, Xiong, Bradbury, and Socher]{wikitext}
Stephen Merity, Caiming Xiong, James Bradbury, and Richard Socher.
\newblock Pointer sentinel mixture models.
\newblock In \emph{5th International Conference on Learning Representations,
  {ICLR} 2017, Toulon, France, April 24-26, 2017, Conference Track
  Proceedings}. OpenReview.net, 2017.

\bibitem[Narayanan et~al.(2021)Narayanan, Shoeybi, Casper, LeGresley, Patwary,
  Korthikanti, Vainbrand, Kashinkunti, Bernauer, Catanzaro, et~al.]{megatron_2}
Deepak Narayanan, Mohammad Shoeybi, Jared Casper, Patrick LeGresley, Mostofa
  Patwary, Vijay~Anand Korthikanti, Dmitri Vainbrand, Prethvi Kashinkunti,
  Julie Bernauer, Bryan Catanzaro, et~al.
\newblock Efficient large-scale language model training on gpu clusters.
\newblock \emph{arXiv preprint arXiv:2104.04473}, 2021.

\bibitem[Paperno et~al.(2016)Paperno, Kruszewski, Lazaridou, Pham, Bernardi,
  Pezzelle, Baroni, Boleda, and Fern{\'a}ndez]{lambada}
Denis Paperno, Germ{\'a}n Kruszewski, Angeliki Lazaridou, Ngoc~Quan Pham,
  Raffaella Bernardi, Sandro Pezzelle, Marco Baroni, Gemma Boleda, and Raquel
  Fern{\'a}ndez.
\newblock The {LAMBADA} dataset: Word prediction requiring a broad discourse
  context.
\newblock In \emph{Proceedings of the 54th Annual Meeting of the Association
  for Computational Linguistics (Volume 1: Long Papers)}, pp.\  1525--1534,
  Berlin, Germany, August 2016. Association for Computational Linguistics.

\bibitem[Patterson et~al.(2021)Patterson, Gonzalez, Le, Liang, Munguia,
  Rothchild, So, Texier, and Dean]{carbon_emission}
David~A. Patterson, Joseph Gonzalez, Quoc~V. Le, Chen Liang, Lluis{-}Miquel
  Munguia, Daniel Rothchild, David~R. So, Maud Texier, and Jeff Dean.
\newblock Carbon emissions and large neural network training.
\newblock \emph{arXiv preprint arXiv:2104.10350}, 2021.

\bibitem[Radford et~al.(2018)Radford, Narasimhan, Salimans, and Sutskever]{gpt}
Alec Radford, Karthik Narasimhan, Tim Salimans, and Ilya Sutskever.
\newblock Improving language understanding by generative pre-training.
\newblock \emph{URL
  https://s3-us-west-2.amazonaws.com/openai-assets/researchcovers/
  languageunsupervised/language understanding paper. pdf}, 2018.

\bibitem[Radford et~al.(2019)Radford, Wu, Child, Luan, Amodei, and
  Sutskever]{gpt2}
Alec Radford, Jeffrey Wu, Rewon Child, David Luan, Dario Amodei, and Ilya
  Sutskever.
\newblock Language models are unsupervised multitask learners, 2019.

\bibitem[Radford et~al.(2021)Radford, Kim, Hallacy, Ramesh, Goh, Agarwal,
  Sastry, Askell, Mishkin, Clark, Krueger, and Sutskever]{clip}
Alec Radford, Jong~Wook Kim, Chris Hallacy, Aditya Ramesh, Gabriel Goh,
  Sandhini Agarwal, Girish Sastry, Amanda Askell, Pamela Mishkin, Jack Clark,
  Gretchen Krueger, and Ilya Sutskever.
\newblock Learning transferable visual models from natural language
  supervision.
\newblock In \emph{Proceedings of the 38th International Conference on Machine
  Learning, {ICML} 2021}, pp.\  8748--8763, 2021.

\bibitem[Raffel et~al.(2020)Raffel, Shazeer, Roberts, Lee, Narang, Matena,
  Zhou, Li, and Liu]{T5}
Colin Raffel, Noam Shazeer, Adam Roberts, Katherine Lee, Sharan Narang, Michael
  Matena, Yanqi Zhou, Wei Li, and Peter~J Liu.
\newblock Exploring the limits of transfer learning with a unified text-to-text
  transformer.
\newblock \emph{Journal of Machine Learning Research}, 21:\penalty0 1--67,
  2020.

\bibitem[Rajbhandari et~al.(2020)Rajbhandari, Rasley, Ruwase, and He]{zero}
Samyam Rajbhandari, Jeff Rasley, Olatunji Ruwase, and Yuxiong He.
\newblock Zero: Memory optimizations toward training trillion parameter models.
\newblock In \emph{SC20: International Conference for High Performance
  Computing, Networking, Storage and Analysis}, pp.\  1--16. IEEE, 2020.

\bibitem[Rajbhandari et~al.(2021)Rajbhandari, Ruwase, Rasley, Smith, and
  He]{zero_infinity}
Samyam Rajbhandari, Olatunji Ruwase, Jeff Rasley, Shaden Smith, and Yuxiong He.
\newblock Zero-infinity: Breaking the gpu memory wall for extreme scale deep
  learning.
\newblock \emph{arXiv preprint arXiv:2104.07857}, 2021.

\bibitem[Ramesh et~al.(2021)Ramesh, Pavlov, Goh, Gray, Voss, Radford, Chen, and
  Sutskever]{dalle}
Aditya Ramesh, Mikhail Pavlov, Gabriel Goh, Scott Gray, Chelsea Voss, Alec
  Radford, Mark Chen, and Ilya Sutskever.
\newblock Zero-shot text-to-image generation.
\newblock \emph{arXiv preprint arXiv:2102.12092}, 2021.

\bibitem[Rasley et~al.(2020)Rasley, Rajbhandari, Ruwase, and He]{deepspeed}
Jeff Rasley, Samyam Rajbhandari, Olatunji Ruwase, and Yuxiong He.
\newblock Deepspeed: System optimizations enable training deep learning models
  with over 100 billion parameters.
\newblock In \emph{Proceedings of the 26th ACM SIGKDD International Conference
  on Knowledge Discovery \& Data Mining}, pp.\  3505--3506, 2020.

\bibitem[Ren et~al.(2021)Ren, Rajbhandari, Aminabadi, Ruwase, Yang, Zhang, Li,
  and He]{zero_offload}
Jie Ren, Samyam Rajbhandari, Reza~Yazdani Aminabadi, Olatunji Ruwase, Shuangyan
  Yang, Minjia Zhang, Dong Li, and Yuxiong He.
\newblock Zero-offload: Democratizing billion-scale model training.
\newblock \emph{arXiv preprint arXiv:2101.06840}, 2021.

\bibitem[Roller et~al.(2021)Roller, Sukhbaatar, Szlam, and Weston]{hash-layers}
Stephen Roller, Sainbayar Sukhbaatar, Arthur Szlam, and Jason Weston.
\newblock Hash layers for large sparse models.
\newblock \emph{arXiv preprint arXiv:2106.04426}, 2021.

\bibitem[Rosset(2020)]{turing-nlg}
Corby Rosset.
\newblock Turing-nlg: A 17-billion parameter language model by microsoft.
\newblock \emph{Microsoft Research Blog}, 2020.

\bibitem[Schwartz et~al.(2020)Schwartz, Dodge, Smith, and Etzioni]{green_ai}
Roy Schwartz, Jesse Dodge, Noah~A. Smith, and Oren Etzioni.
\newblock Green {AI}.
\newblock \emph{Commun. {ACM}}, 63\penalty0 (12):\penalty0 54--63, 2020.

\bibitem[Shazeer et~al.(2017)Shazeer, Mirhoseini, Maziarz, Davis, Le, Hinton,
  and Dean]{moe}
Noam Shazeer, Azalia Mirhoseini, Krzysztof Maziarz, Andy Davis, Quoc~V. Le,
  Geoffrey~E. Hinton, and Jeff Dean.
\newblock Outrageously large neural networks: The sparsely-gated
  mixture-of-experts layer.
\newblock In \emph{5th International Conference on Learning Representations,
  {ICLR} 2017}, 2017.

\bibitem[Shazeer et~al.(2018)Shazeer, Cheng, Parmar, Tran, Vaswani,
  Koanantakool, Hawkins, Lee, Hong, Young, Sepassi, and Hechtman]{mesh}
Noam Shazeer, Youlong Cheng, Niki Parmar, Dustin Tran, Ashish Vaswani, Penporn
  Koanantakool, Peter Hawkins, HyoukJoong Lee, Mingsheng Hong, Cliff Young,
  Ryan Sepassi, and Blake~A. Hechtman.
\newblock Mesh-tensorflow: Deep learning for supercomputers.
\newblock In \emph{Advances in Neural Information Processing Systems 31: Annual
  Conference on Neural Information Processing Systems 2018, NeurIPS 2018}, pp.\
   10435--10444, 2018.

\bibitem[Shoeybi et~al.(2019)Shoeybi, Patwary, Puri, LeGresley, Casper, and
  Catanzaro]{megatron}
Mohammad Shoeybi, Mostofa Patwary, Raul Puri, Patrick LeGresley, Jared Casper,
  and Bryan Catanzaro.
\newblock Megatron-lm: Training multi-billion parameter language models using
  model parallelism.
\newblock \emph{arXiv preprint arXiv:1909.08053}, 2019.

\bibitem[Sun et~al.(2021)Sun, Wang, Feng, Ding, Pang, Shang, Liu, Chen, Zhao,
  Lu, et~al.]{ernie3}
Yu~Sun, Shuohuan Wang, Shikun Feng, Siyu Ding, Chao Pang, Junyuan Shang,
  Jiaxiang Liu, Xuyi Chen, Yanbin Zhao, Yuxiang Lu, et~al.
\newblock Ernie 3.0: Large-scale knowledge enhanced pre-training for language
  understanding and generation.
\newblock \emph{arXiv preprint arXiv:2107.02137}, 2021.

\bibitem[Vaswani et~al.(2017)Vaswani, Shazeer, Parmar, Uszkoreit, Jones, Gomez,
  Kaiser, and Polosukhin]{transformer}
Ashish Vaswani, Noam Shazeer, Niki Parmar, Jakob Uszkoreit, Llion Jones,
  Aidan~N. Gomez, Lukasz Kaiser, and Illia Polosukhin.
\newblock Attention is all you need.
\newblock In Isabelle Guyon, Ulrike von Luxburg, Samy Bengio, Hanna~M. Wallach,
  Rob Fergus, S.~V.~N. Vishwanathan, and Roman Garnett (eds.), \emph{Advances
  in Neural Information Processing Systems 30: Annual Conference on Neural
  Information Processing Systems 2017, December 4-9, 2017, Long Beach, CA,
  {USA}}, pp.\  5998--6008, 2017.

\bibitem[Xiong et~al.(2020)Xiong, Yang, He, Zheng, Zheng, Xing, Zhang, Lan,
  Wang, and Liu]{prenorm_analysis}
Ruibin Xiong, Yunchang Yang, Di~He, Kai Zheng, Shuxin Zheng, Chen Xing,
  Huishuai Zhang, Yanyan Lan, Liwei Wang, and Tie{-}Yan Liu.
\newblock On layer normalization in the transformer architecture.
\newblock In \emph{Proceedings of the 37th International Conference on Machine
  Learning, {ICML} 2020, 13-18 July 2020, Virtual Event}, volume 119 of
  \emph{Proceedings of Machine Learning Research}, pp.\  10524--10533. {PMLR},
  2020.

\bibitem[Xue et~al.(2021)Xue, Constant, Roberts, Kale, Al-Rfou, Siddhant,
  Barua, and Raffel]{mt5}
Linting Xue, Noah Constant, Adam Roberts, Mihir Kale, Rami Al-Rfou, Aditya
  Siddhant, Aditya Barua, and Colin Raffel.
\newblock m{T}5: A massively multilingual pre-trained text-to-text transformer.
\newblock In \emph{Proceedings of the 2021 Conference of the North American
  Chapter of the Association for Computational Linguistics: Human Language
  Technologies}, pp.\  483--498, Online, June 2021. Association for
  Computational Linguistics.

\bibitem[Yang et~al.(2021)Yang, Lin, Men, Zhou, Jiang, Jia, Wang, Zhang, Wang,
  Li, et~al.]{m6-t}
An~Yang, Junyang Lin, Rui Men, Chang Zhou, Le~Jiang, Xianyan Jia, Ang Wang, Jie
  Zhang, Jiamang Wang, Yong Li, et~al.
\newblock M6-t: Exploring sparse expert models and beyond.
\newblock \emph{arXiv preprint arXiv:2105.15082}, 2021.

\bibitem[You et~al.(2020)You, Li, Reddi, Hseu, Kumar, Bhojanapalli, Song,
  Demmel, Keutzer, and Hsieh]{lamb}
Yang You, Jing Li, Sashank~J. Reddi, Jonathan Hseu, Sanjiv Kumar, Srinadh
  Bhojanapalli, Xiaodan Song, James Demmel, Kurt Keutzer, and Cho{-}Jui Hsieh.
\newblock Large batch optimization for deep learning: Training {BERT} in 76
  minutes.
\newblock In \emph{8th International Conference on Learning Representations,
  {ICLR} 2020, Addis Ababa, Ethiopia, April 26-30, 2020}. OpenReview.net, 2020.

\bibitem[Zeng et~al.(2021)Zeng, Ren, Su, Wang, Liao, Wang, Jiang, Yang, Wang,
  Zhang, et~al.]{pangu-alpha}
Wei Zeng, Xiaozhe Ren, Teng Su, Hui Wang, Yi~Liao, Zhiwei Wang, Xin Jiang,
  ZhenZhang Yang, Kaisheng Wang, Xiaoda Zhang, et~al.
\newblock Pangu-$\alpha$: Large-scale autoregressive pretrained chinese
  language models with auto-parallel computation.
\newblock \emph{arXiv preprint arXiv:2104.12369}, 2021.

\bibitem[Zhang \& He(2020)Zhang and He]{pld}
Minjia Zhang and Yuxiong He.
\newblock Accelerating training of transformer-based language models with
  progressive layer dropping.
\newblock In Hugo Larochelle, Marc'Aurelio Ranzato, Raia Hadsell,
  Maria{-}Florina Balcan, and Hsuan{-}Tien Lin (eds.), \emph{Advances in Neural
  Information Processing Systems 33: Annual Conference on Neural Information
  Processing Systems 2020, NeurIPS 2020, December 6-12, 2020, virtual}, 2020.

\bibitem[Zhang et~al.(2021{\natexlab{a}})Zhang, Gu, Han, Chen, Xiao, Sun, Yao,
  Qi, Guan, Ke, et~al.]{cpm-2}
Zhengyan Zhang, Yuxian Gu, Xu~Han, Shengqi Chen, Chaojun Xiao, Zhenbo Sun, Yuan
  Yao, Fanchao Qi, Jian Guan, Pei Ke, et~al.
\newblock Cpm-2: Large-scale cost-effective pre-trained language models.
\newblock \emph{arXiv preprint arXiv:2106.10715}, 2021{\natexlab{a}}.

\bibitem[Zhang et~al.(2021{\natexlab{b}})Zhang, Han, Zhou, Ke, Gu, Ye, Qin, Su,
  Ji, Guan, et~al.]{cpm}
Zhengyan Zhang, Xu~Han, Hao Zhou, Pei Ke, Yuxian Gu, Deming Ye, Yujia Qin,
  Yusheng Su, Haozhe Ji, Jian Guan, et~al.
\newblock Cpm: A large-scale generative chinese pre-trained language model.
\newblock \emph{AI Open}, 2:\penalty0 93--99, 2021{\natexlab{b}}.

\bibitem[Zhang et~al.(2021{\natexlab{c}})Zhang, Ma, Zhou, Men, Li, Ding, Tang,
  Zhou, and Yang]{ufc-bert}
Zhu Zhang, Jianxin Ma, Chang Zhou, Rui Men, Zhikang Li, Ming Ding, Jie Tang,
  Jingren Zhou, and Hongxia Yang.
\newblock M6-ufc: Unifying multi-modal controls for conditional image
  synthesis.
\newblock \emph{arXiv preprint arXiv:2105.14211}, 2021{\natexlab{c}}.

\bibitem[Zhu et~al.(2015)Zhu, Kiros, Zemel, Salakhutdinov, Urtasun, Torralba,
  and Fidler]{bookcorpus}
Yukun Zhu, Ryan Kiros, Richard~S. Zemel, Ruslan Salakhutdinov, Raquel Urtasun,
  Antonio Torralba, and Sanja Fidler.
\newblock Aligning books and movies: Towards story-like visual explanations by
  watching movies and reading books.
\newblock In \emph{2015 {IEEE} International Conference on Computer Vision,
  {ICCV} 2015, Santiago, Chile, December 7-13, 2015}, pp.\  19--27. {IEEE}
  Computer Society, 2015.

\end{thebibliography}
\bibliographystyle{iclr2022_conference}


\end{document}